%% file: main.tex
\documentclass[10pt,twocolumn,letterpaper]{article}

\usepackage{wacv}
\makeatletter
\@namedef{ver@everyshi.sty}{}
\makeatother

\usepackage{times}
\usepackage{epsfig}
\usepackage{graphicx}
\usepackage{amsmath}
\usepackage{amssymb}

\usepackage{import}
\usepackage{pgfplots}
\usepgfplotslibrary{groupplots}
\usepackage{subcaption}
\usepackage{multirow}
\pgfplotsset{compat=1.9}

\def\wacvPaperID{8} 

\wacvfinalcopy 

\ifwacvfinal
\def\assignedStartPage{1} 
\fi

\ifwacvfinal
\usepackage[breaklinks=true,bookmarks=false]{hyperref}
\else
\usepackage[pagebackref=true,breaklinks=true,colorlinks,bookmarks=false]{hyperref}
\fi

\ifwacvfinal
\else
\fi

\begin{document}

\title{Joint Visual-Temporal Embedding for Unsupervised Learning of Actions in Untrimmed Sequences}

\author{\vspace{-1.8mm}\parbox{16cm}{\centering
    {\large Rosaura G. VidalMata$^1$, 
    Walter J. Scheirer$^1$, Anna Kukleva\textsuperscript{2}, David Cox\textsuperscript{3} and Hilde Kuehne\textsuperscript{3}}\\
    {\normalsize
    $^1$ University of Notre Dame, IN, USA  $^2$ Max-Planck-Institute for Informatics, Saarbrucken, Germany $^3$ MIT-IBM Waston AI Lab, MA, USA\\
        \tt\small \{rvidalma, walter.scheirer\}@nd.edu, 
        \tt\small akukleva@mpi-inf.mpg.de,
        \tt\small \{david.d.cox, kuehne\}@ibm.com
    }   
    }
    }


\maketitle

\begin{abstract}
   Understanding the structure of complex activities in untrimmed videos is a challenging task in the area of action recognition. One problem here is that this task usually requires a large amount of hand-annotated minute- or even hour-long video data, but annotating such data is very time consuming and can not easily be automated or scaled. To address this problem, this paper proposes an approach for the unsupervised learning of actions in untrimmed video sequences based on a joint visual-temporal embedding space.  To this end, we combine a visual embedding based on a predictive U-Net architecture with a temporal continuous function. The resulting representation space allows detecting relevant action clusters based on their visual as well as their temporal appearance. The proposed method is evaluated on three standard benchmark datasets, Breakfast Actions, INRIA YouTube Instructional Videos, and 50 Salads. We show that the proposed approach is able to provide a meaningful visual and temporal embedding out of the visual cues present in contiguous video frames and is suitable for the task of unsupervised temporal segmentation of actions.
\end{abstract}

\vspace{-1em}
\section{Introduction}
Research shows that humans usually understand complex activities through the ongoing temporal segmentation of perceived inputs into meaningful segments~\cite{zacks2007event}. Nevertheless, replicating this behavior in fully automated systems is a challenging problem as it requires identifying the meaningful steps in a given task and how do they logically relate to each other. Fully supervised systems have been proposed to address this, but they rely on large amounts of training data. Manually annotating this data is especially expensive for temporal video segmentation as this task usually requires a dense frame-based annotation. Weakly supervised approaches attempt to alleviate this by incorporating the use of additional sources of human-generated information such as speech or video captions~\cite{Alayrac_2016_CVPR,malmaud2015s}. However, differences between the alignment of additional modalities like audio, subtitles, or descriptive meta-data to the video frames might prove challenging for some of these approaches~\cite{DBLP:journals/corr/SenerZWSS16}.

\begin{figure}[t]
    \centering 
    \begin{subfigure}[]{\linewidth}
    \centering
    \caption{Breakfast Actions (BF)}\label{fig:teasercof}
    \includegraphics[width=0.8\linewidth]{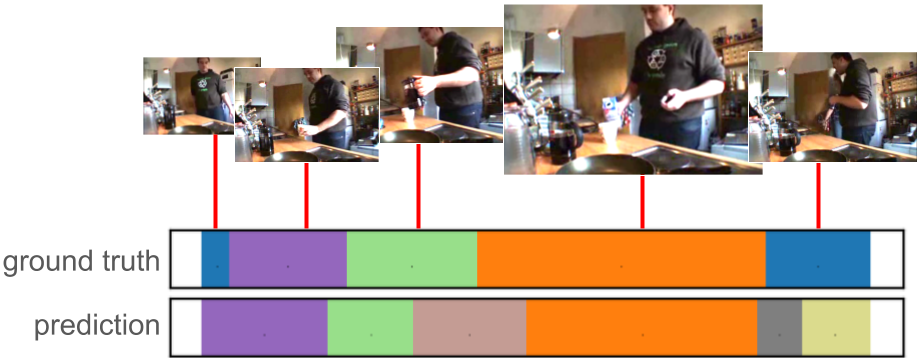}\vspace{3mm}
    \end{subfigure} 
    \begin{subfigure}[]{\linewidth}
    \centering
    \caption{INRIA YouTube Instructional Videos (YTI)}\label{fig:teasertire}
    \includegraphics[width=0.8\linewidth]{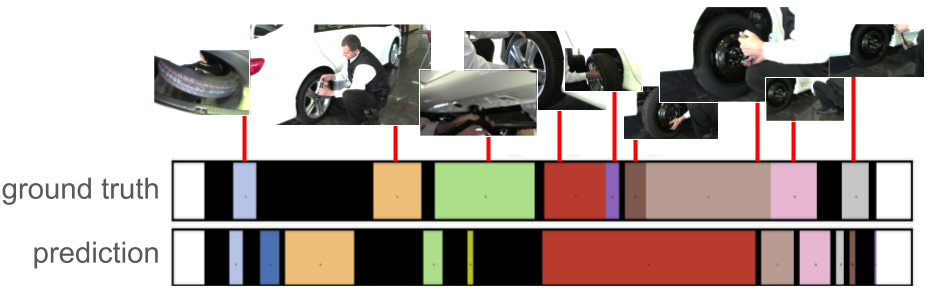}
    \end{subfigure}
    \caption{Temporal segmentation of two videos from the Breakfast Actions~\cite{Kuehne_2014_CVPR} and INRIA YouTube Instructional Videos (YTI)~\cite{Alayrac_2016_CVPR} Datasets. The black segments in the YTI video correspond to background frames whose content is not associated with a sub-activity relevant to the video task. Our approach maintains the logical ordering of sub-activities and a good estimate on their start and duration.\vspace{-1em}}\label{fig:teaser}
\end{figure}

Therefore, other methods have been proposed to tackle the challenge of training such models without dense human-generated labels, spanning from learning with weak or sparse annotation~\cite{huang2016connectionist,kuehne2019hybird,li2019weakly} to completely unsupervised learning of temporal action segmentation~\cite{bojanowski2014weakly,sener2018unsupervised,Kukleva_2019_CVPR,Aakur_2019_CVPR}. The work described in this paper deals with the latter problem, the unsupervised learning of action segments from unlabeled video data, which can be framed as the task of unsupervised temporal action segmentation. This follows the idea that, given a set of videos all capturing the same activity, it should be possible to identify temporal segments with similar sub-actions across all videos.

Previous work~\cite{bojanowski2014weakly,sener2018unsupervised,Kukleva_2019_CVPR} has shown that temporal appearance can play a critical role in this task. For instance, in videos showing how to make pancakes the process of cracking eggs should not only look visually similar across different videos but would generally occur before other tasks such as mixing the eggs into the batter or pouring the batter onto the griddle. So, it can be assumed that actions, at least in task-oriented videos, not only share certain visual features but also occur in a similar temporal space. Recent approaches usually incorporate this property by learning a strong temporal regularization~\cite{sener2018unsupervised,Kukleva_2019_CVPR}, which helps to find clusters over time but may result in a lower ability to identify segments based on their visual representation.

To address this problem, this paper proposes a joint visual-temporal learning pipeline that combines the advantages of current temporal embedding systems with a visual embedding based on a combination of predictive visual and temporal learning tasks. To this end, we combine a state-of-the-art temporal embedding system~\cite{Kukleva_2019_CVPR}, designed to estimate the relative timestamp of a given video frame, with a visual encoder-decoder pipeline that is trained on a combination of visual and temporal losses.

The intuition behind this is that the embedding should not only reconstruct the plain input signal, but it should also find a reconstruction that allows for a better estimation of the respective timestamp and, thus, a better temporal reconstruction. To prevent the learning of weak visual cues, we shift the output of the visual encoder by a number of frames, which turns it into a visual prediction architecture, similar to other self-supervised models~\cite{mathieu2015deep, NIPS2017_7028}. Combined with the temporal loss, the model predicts a frame-representation that is optimized to give the best timestamp prediction in the temporal embedding framework. The resulting embedding space thus captures visual and temporal representations of each individual frame.

We evaluate the proposed system on three challenging standard benchmark datasets, the Breakfast dataset~\cite{Kuehne_2014_CVPR}, the INRIA YouTube Instructional Videos (YTI) dataset~\cite{bojanowski2014weakly}, and the 50 Salads dataset~\cite{stein2013combining}, achieving good temporal segmentation compared to state-of-the-art approaches. Figure~\ref{fig:teaser} shows qualitative examples which demonstrate that the proposed architecture is able to adapt well to a diverse set of action tasks.

\vspace{-0.75em}
\section{Related Work}\vspace{-0.5em}

\subsection{Unsupervised learning of temporal sequences}\vspace{-0.25em}

While there has been plenty of work done in the area of action segmentation in video, a vast majority of the approaches are fully supervised methods relying on frame annotations~\cite{huang2016connectionist, ding2018weakly, kuehne2016end, richard2017weakly}, or weakly supervised approaches that use some form of metadata~\cite{li2019weakly,chang2019d3tw,richard2018nnviterbi}. Supervised models achieve high-quality temporal segmentations but their training is heavily dependent on vast amounts of good quality hand-annotated minute-or even hour-long video data which can become prohibitive in most real-life scenarios as this process can be time-consuming and can not easily be automated or scaled. 

To overcome the dependence on labeled data, unsupervised methods have seen increased interest recently. Chen~\etal~\cite{chen2020action} propose a domain adaptation approach based on self-supervised auxiliary tasks (SSTDA) that is able to obtain state-of-the-art performance with a reduced amount of labeled data (they show that even when using $65\%$ of the labeled data their method still produces state-of-the-art results). One of the first unsupervised methods for action segmentation is the one proposed by Bojanovski \etal~\cite{bojanowski2014weakly} based on a Frank-Wolfe optimization algorithm. Sener \etal~\cite{sener2018unsupervised} later proposed the modeling of the temporal structure of sub-activities using a combination of Generalized Mallows Model (GMM) sampling and the estimation of the action length calculated using the frame distribution to estimate the action segmentation in complex action videos. Following a similar segmentation pipeline, Kukleva \etal~\cite{Kukleva_2019_CVPR} proposed instead a combination of temporal encoding generated using a frame timestamp prediction network and Viterbi decoding for consistent frame-to-cluster assignment. Another interesting take on the problem is further proposed by Aakar \etal~\cite{Aakur_2019_CVPR}. Different from the methods so far, the idea here is not to cluster the feature space to identify actions, but to detect action boundaries in an unsupervised way by learning a predictive framework which uses the difference between observed and predicted frame features as a means to determine event boundaries. As such the approach is focusing on a per video-based segmentation instead of the identification of action classes across multiple videos.

\begin{figure*}[t]
    \centering
    \includegraphics[width=0.85\linewidth]{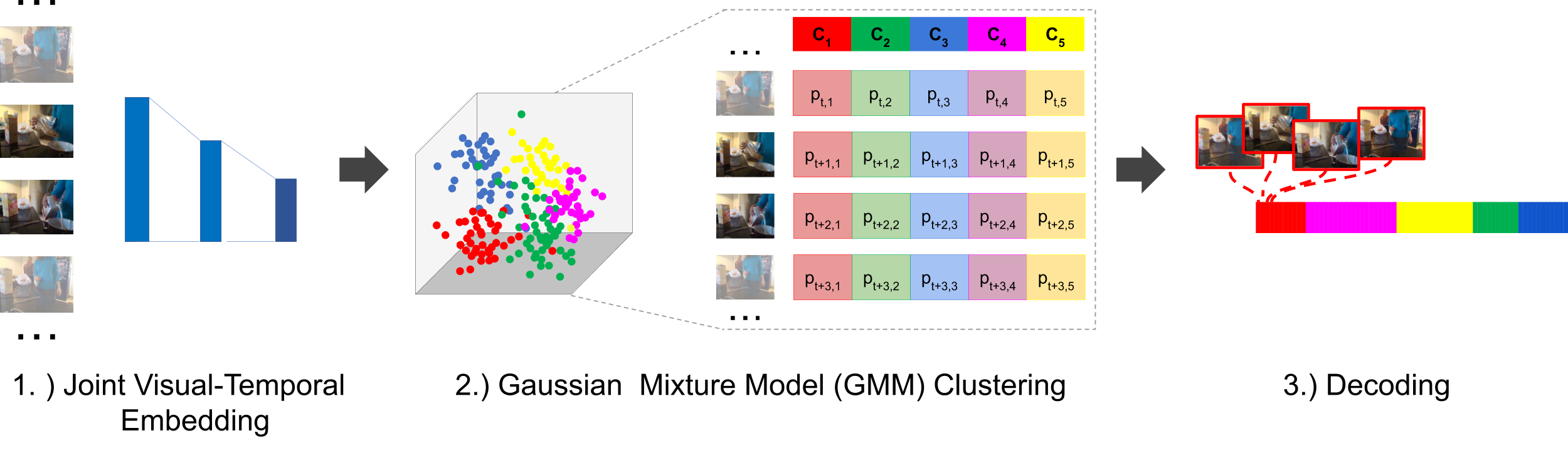} \vspace{-0.75em}
\caption{Joint visual-temporal training pipeline. If the feature embedding has a good representation of the visual and temporal attributes of each frame, the frames that cluster together will have similar temporal locations and share visual attributes that represent a given sub-activity.\vspace{-0.75em}}
\label{fig:pipeline}
\end{figure*}

\subsection{Unsupervised and self-supervised learning of visual representations} \vspace{-0.5em}

Various approaches have been proposed for the learning of visual representations without labels~\cite{Jing2019survey}. Particularly, in the case of learning video representations in an unsupervised way, various approaches make use of temporal properties of the data \eg, in the form of shuffling~\cite{MisraZH16Shuffle}, or similar to the idea proposed in this paper, temporal prediction~\cite{srivastava2015Unsupervised,NIPS2017_7028}. Temporal prediction has been established as a way to achieve a deeper understanding of the data as it requires an implicit understanding of the structure of the observed visual features and the rules they follow while they change over time~\cite{srivastava2015Unsupervised,DBLP:journals/corr/LotterKC16,DBLP:journals/corr/WhitneyCKT16}. However, it has been pointed out that the use of traditional losses does not translate well for video frame prediction. Srivastava \etal~\cite{srivastava2015Unsupervised} observed that predictive models trained solely with an MSE loss have a tendency of blurring regions with uncertainty. 

This has given way to the introduction of promising alternatives such as the adversarial loss present in Generative Adversarial Networks (GANs)~\cite{goodfellow2014generative} and Conditional GANs~\cite{mirza2014conditional}, which have led to significant advances in the performance of video prediction~\cite{NIPS2017_7028,DBLP:journals/corr/LotterKC15,mathieu2015deep}. GAN-based methods have incorporated the use of U-Net architectures~\cite{DBLP:journals/corr/RonnebergerFB15} into their generator module given their good performance in image-to-image translation~\cite{isola2017image,zhu2017toward} and image segmentation~\cite{DBLP:journals/corr/RonnebergerFB15,badrinarayanan2017segnet,long2015fully} tasks. Our proposed approach does not focus on learning feature representations as is the case in those methods but instead borrows other ideas from the field to improve the joint visual-temporal learning of an embedding space for unsupervised actions recognition.

\begin{figure*}[t]
    \centering
    \begin{subfigure}{0.3\textwidth}
    \includegraphics[width=\linewidth]{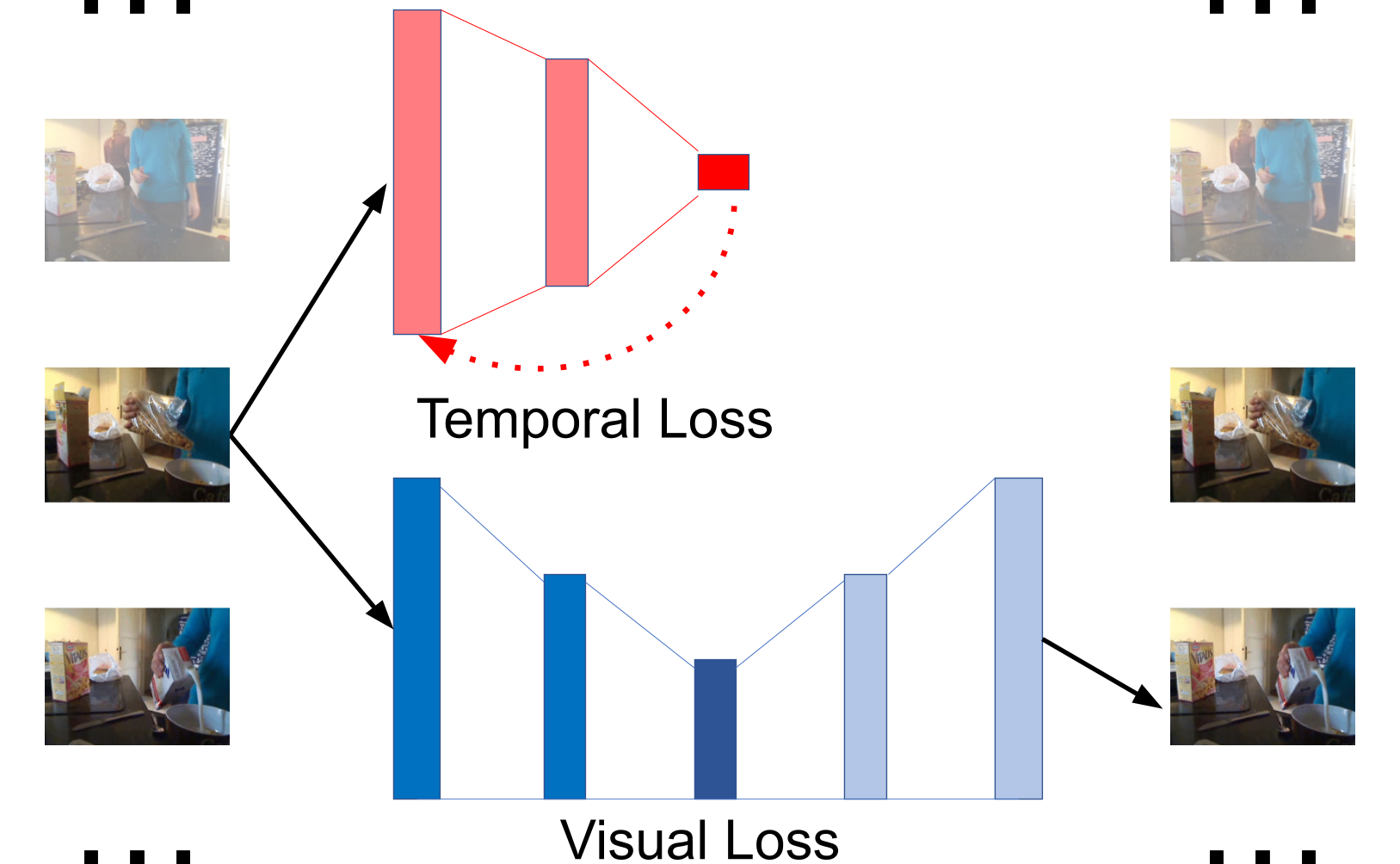}
    \caption{Stage 1}\label{fig:train_stage1}
    \end{subfigure} \hspace{3em}
    \begin{subfigure}{0.6\textwidth}
    \includegraphics[width=0.5\linewidth]{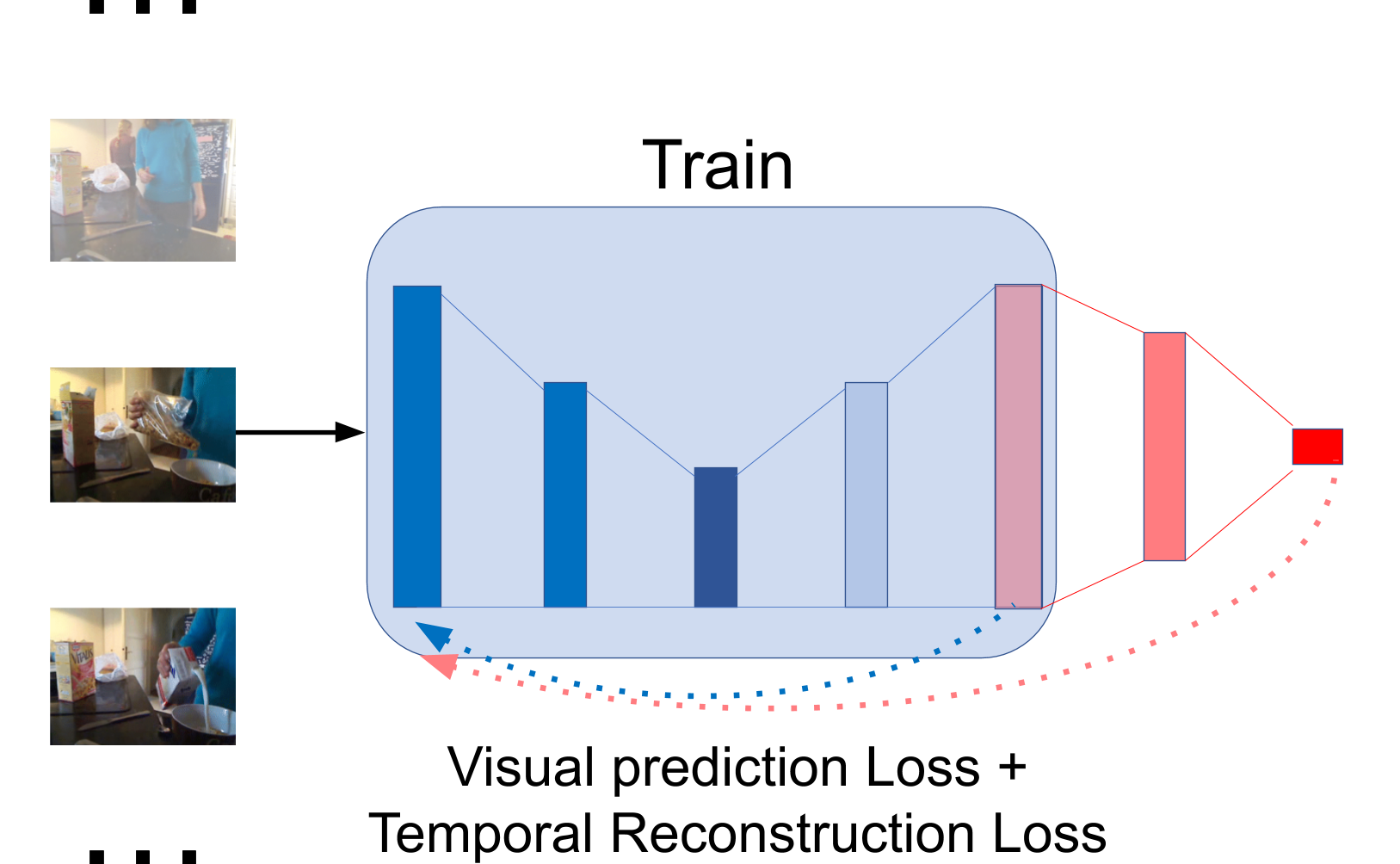}
    \hspace{1em}
    \includegraphics[width=0.5\linewidth]{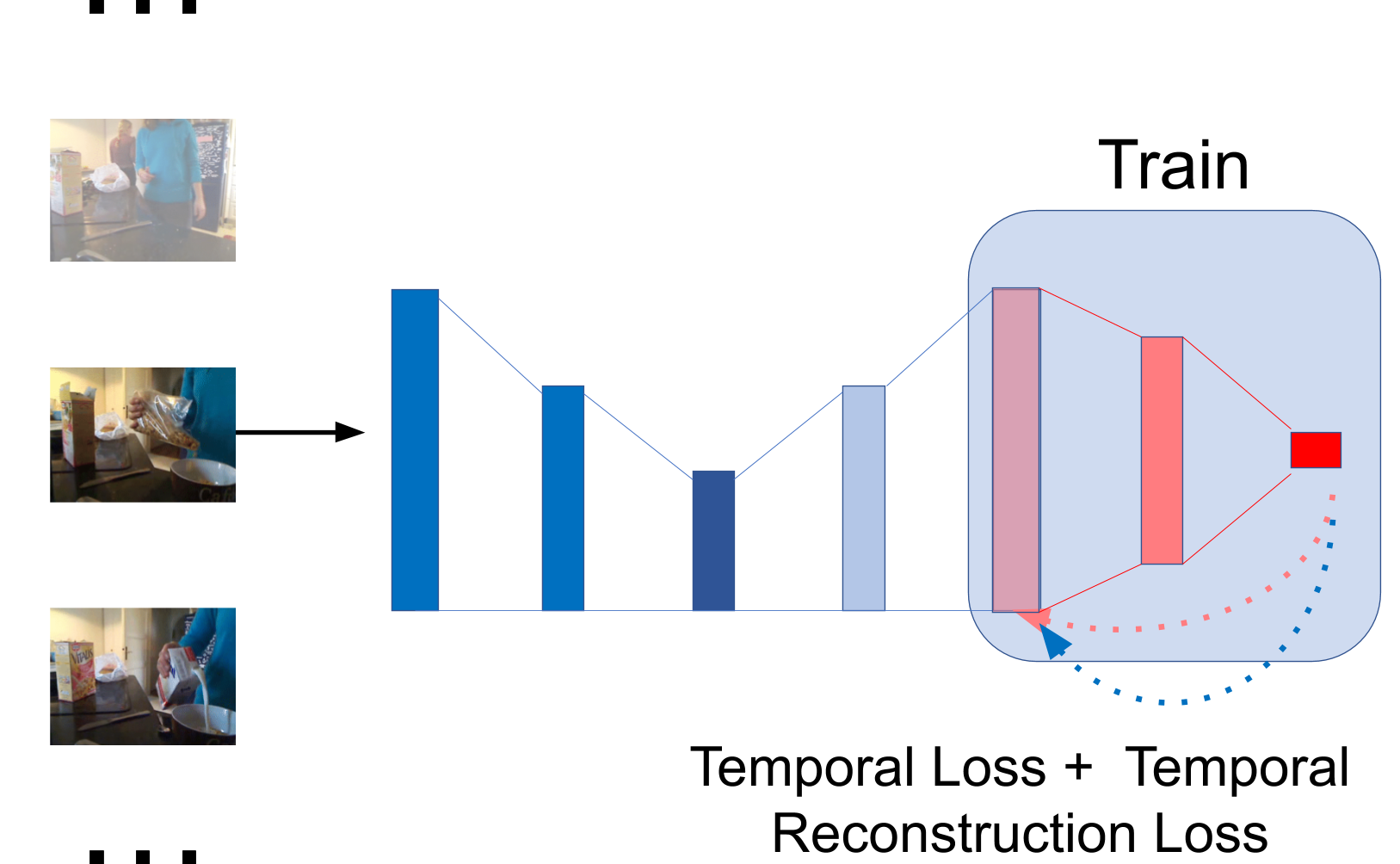}
    \caption{Stage 2}\label{fig:train_stage2}
    \end{subfigure}
\caption{Two-stage visual-embedding pipeline: \ref{fig:train_stage1}) Visual+Temporal embedding: next frame prediction U-Net to generate a visual-temporal embedding (output of the last down-sampling layer, denoted by a red arrow), \ref{fig:train_stage2}) Temporal discriminator: timestamp predictor MLP used to identify the loss of temporal information in the frames predicted by stage 1.\vspace{-0.75em}}
\label{fig:train}
\end{figure*}

\vspace{-0.5em}
\section{System Description} \label{systemdescription}\vspace{-0.25em}

Given a collection of complex activity videos $D_{A}=\{v_{i}\}_{i=1}^V$ with $V$ videos belonging all to the same activity class $A$, we want to learn $k$ action classes $C_{k}$ that appear in this activity, as well as the action class label of all the frames ($N_{v_i}$) of each video $v_{i}=\{f_{n}\}_{n=1}^{N_{v_i}}$ . Note that the frame input used here is a pre-extracted one-dimensional feature vector, that can be derived from hand-crafted feature representations or from a network that has been trained in an unsupervised way.

To learn an embedding space for this input, we propose a two-stage pipeline (see Figure~\ref{fig:train}). The first stage (\ref{fig:train_stage1}) consists of the independent training of two models: one encoding a visual embedding and the second encoding a temporal embedding only. We pre-train both models separately and then join them in the second stage (\ref{fig:train_stage2}), where the temporal model becomes a discriminator for the frames generated by the visual model. We update the components of this model in an alternate fashion, keeping the visual model constant during the training phase of the temporal model in order to allow it to identify the loss of valuable temporal information from the reconstructed frame. Similarly, we then keep the temporal model constant while the visual model is training to improve the results of its outputs when they are evaluated by the temporal discriminator.

This joint training allows our system to learn an embedding space that not only captures visual properties by predicting the next frame but also ensures that we retain enough temporal information to estimate a consistent timestamp for the output. The embedding space of the model is then used for clustering to form the respective classes and segment the videos accordingly. We illustrate our clustering and segmentation pipeline during testing in Figure~\ref{fig:pipeline}. We discuss each step of the pipeline in the following sections.

\vspace{-0.25em}
\subsection{Stage 1: Disjoint training of the Visual and Temporal Models}\label{sys:s1:visual} \vspace{-0.5em}
\subsubsection{Visual Embedding Model}  \vspace{-0.25em}

The visual embedding model is designed to learn a low dimensional embedding of the frame-based input vector. 
For this, we make use of an encoder-decoder structure to model a frame prediction framework which, using as a prior the visual features of the frame at time $t$ (frame $f_{t}$) learns to predict the features at a future time $t+s$ ($f_{t+s}$). 

As is shown in Figure~\ref{fig:train_stage1}, our visual model (displayed in blue) consists of three down-sampling layers followed by three up-sampling layers with skip connections between the down-sampling and up-sampling layers to preserve fine-grained details from being discarded by the encoding process. As such, we are able to suppress visual noise and find visual cues that are more relevant to the respective task while ensuring that the encoded embedding $x_{t}$ maintains enough relevant information in order to reconstruct a viable future frame in the video sequence. This linear U-Net is then used during testing to generate an abstract representation of the frame's features $x_{t}$ (generated by the last down-sampling layer), which is then used as the embedding of the frame $f_{t}$ during the later clustering process (see Figure~\ref{fig:pipeline}).

This visual embedding model learns to generate accurate future frame predictions by minimizing the mean squared difference between the predicted and the real frames:
\begin{equation}Loss_{vis} = \frac{1}{N-s}  \sum_{t=1}^{N-s} (f_{t+s} - \hat{f}_{t+s})^{2} \label{f:visual_loss}\end{equation} For ease of notation, we use $N$ as the number of all frames of all videos $V$ belonging to the same activity class $A$, and $s$ as the prediction time step. That is, for $s=5$ given an input frame captured at timestamp $t=0$ ($f_{t=0}$) the network will return a predicted frame $s$ frames into the future ($\hat{f}_{t=5}$). For our experiments we used $s=5$ as the prediction goal of our visual embedding model. However, we study the impact of different step sizes in Section~\ref{sec:ablation}.

\vspace{-0.5em}
\subsubsection{Temporal Embedding Model}

The temporal embedding model (displayed in red in Figure~\ref{fig:train_stage1}) is designed to extract valuable temporal information out of a given frame. For this we implemented a three-layer Multilayer Perceptron (MLP) with the learning goal of predicting the relative timestamp $t$ of a given frame (where $t = frameIdx / N$). The MLP receives as input a video frame $f_{t}$ captured at a timestamp $t$ and provides a prediction for the relative timestamp of the frame $T(f_{t})$. As such, to ensure the MLP has a working knowledge on the temporal structure of the data, it must minimize the difference between the timestamp prediction $T(f_{n})$ it provides for a given $f_{t}$  frame, and the actual timestamp $t$ of such frame: \begin{equation} Loss_{temp} = \frac{1}{N}  \sum_{t=1}^{N} (T(f_{t}) - t)^{2} \label{f:real_loss} \end{equation}

\subsection{Stage 2: Joint visual-temporal training}

In this stage we joined our visual and temporal models in a single framework, as such, our final architecture is composed of (1) a frame predicting U-Net that is able to retain enough visual and temporal features in its predictions, and (2) a timestamp predictor, trained to recognize any discrepancy between the temporal quality of the frame predicted by the U-Net $\hat{f}_{t+s}$ and the observed frame $f_{t+s}$.
\vspace{-0.5em}
\subsubsection{Joint Visual-Temporal Embedding Model}
The learning objective of the U-Net during stage 1 was to predict a frame that seemed visually plausible (Equation~\ref{f:visual_loss}), this helps to ensure that the valuable visual cues in the frame embedding are retained. In the second stage of our training, we build on this objective to ensure that the temporal cues of the frame are also preserved. For this we use our temporal model: the timestamp predictor evaluates the quality of the temporal features encoded into our embedding by measuring the difference between the predicted timestamp of the U-Net's output $T(\hat{f}_{t+s})$ and the predicted timestamp of the observed frame $(T(f_{t+s})$, allowing us to measure the temporal reconstruction loss of our generated frames: \begin{equation} Loss_{trec} = \frac{1}{N-s}  \sum_{t=1}^{N-s} (T(f_{t+s}) - T(\hat{f}_{t+s}))^{2} \label{f:temporal_loss} \end{equation}

We incorporate this temporal reconstruction loss into the U-Net's learning objective: 
\begin{equation} Loss_{joint} = Loss_{vis} + Loss_{trec} \label{f:unet_loss} \end{equation} This joint loss (Equation~\ref{f:unet_loss}) ensures that our embedding model maintains a good balance of visual and temporal cues into its embedding. Its learning objective is now minimizing both the visual and temporal reconstruction losses.

\subsubsection{Temporal Discriminator}

The goal of our temporal embedding model is now to serve as a discriminator for the reconstructed frames generated by our Joint Visual-Temporal Embedding Model. For this, our temporal discriminator must be able to identify the loss of the ``temporal quality" between the frame predicted by the U-Net $\hat{f}_{t+s}$ and the ground-truth frame $f_{t+s}$. Given a frame $f_{t}$ we consider that the U-Net's embedding $x_{t}$ has led to the loss of temporal cues if the predicted timestamp $T(\hat{f}_{t+s})$ (estimated by the temporal embedding model) of the U-Net's output $\hat{f}_{t+s}$ is not similar to the timestamp of the ground-truth frame $f_{t+s}$. The temporal quality ($TQual$) of the U-Net embedding can then be measured as follows:
\begin{equation} TQual = 1 - \left (\frac{1}{N-s}  \sum_{t=1}^{N-s}  T(\hat{f}_{t+s}) - (t+s))^{2}  \right )\label{f:fake_loss} \end{equation}

We can then use the estimation of the temporal quality loss as a means for the MLP to discriminate between a real video frame $f_{t+s}$ and a low-quality estimation $\hat{f}_{t+s}$.  
As a result, the timestamp predictor can learn to discriminate between the temporal differences of the U-Net's prediction and the ground-truth. The loss of the MLP temporal discriminator is then evaluated as follows:
\begin{equation} Loss_{MLP} =  \frac{ \frac{1}{N}  \sum_{t=1}^{N} (T(f_{t}) - (t))^{2} }{TQual} \label{f:mlp_loss} \end{equation} In the above formulation, $T(f_{i})$ represents the output of the timestamp predictor given an input frame $f_{i}$. 

\subsection{Clustering and decoding} \label{sec:clusndecod} 

For all further processing, we follow the protocol of~\cite{Kukleva_2019_CVPR,sener2018unsupervised,bojanowski2014weakly}.  
Once we have the temporally enhanced embedding, we cluster the embedded features of all videos into $k$ clusters. For this, we employ a single Gaussian Mixture Model to model all the embedded features of a given task. To fit them into $k$ clusters, we then use the logarithm of the probability of each frame belonging to each of the clusters.  As such, for every video frame $n$, we have a corresponding $k$ dimensional vector with each of the frame's scores for each of the clusters (see step 2 in Figure~\ref{fig:pipeline}).

A naive approach for segmentation would be using the scores obtained by the clustering method in order to assign a sub-activity to each video frame. This, however, generates a non-homogeneous segmentation with few groups of continuous frames being assigned to the same sub-activity cluster. We use a modified version of the frame labeling method presented in~\cite{Kukleva_2019_CVPR}, Viterbi decoding with length model as proposed by~\cite{kuehne2019hybird} to alleviate this effect. We evaluate the probability of each frame $n$ belonging to a cluster $c_{x}$ with respect to the probabilities of the neighboring frames and seek to maximize the probability of the sequence following a fixed cluster order. The cluster order is determined by the mean time-stamp of each cluster.

A fixed cluster ordering ${c_{1}, c_{2}, ..., c_{i}, c_{j},..., c_{k}}$, would constrain the possible clusters a frame $f_{t}$ can be assigned to. Frame $f_{t}$ could belong to either the same cluster $c_{i}$ as it preceding sampled frame $f_{t-\gamma}$ -where $\gamma$ is the frame sampling size (in the method introduced by~\cite{Kukleva_2019_CVPR} $\gamma = 1$) or to the next cluster $c_{j}$ in the predetermined ordering.

\vspace{-0.5em}
\section{Experimental Evaluation}\vspace{-0.5em}

\subsection{Datasets} \label{evaluation:dataset}
We evaluate our method using three datasets: Breakfast Actions dataset (BF)~\cite{Kuehne_2014_CVPR}, INRIA YouTube Instructional Videos (YTI)~\cite{Alayrac_2016_CVPR}, and 50 Salads (50S)~\cite{stein2013combining}. We are using the reduced Fisher Vector features as proposed by~\cite{kuehne2016end} and used by~\cite{Kukleva_2019_CVPR,sener2018unsupervised,peng2014action} for the evaluation of the three datasets. 

\begin{figure*}[t]
    \centering
    \begin{subfigure}{0.33\textwidth}
    \includegraphics[width=\linewidth]{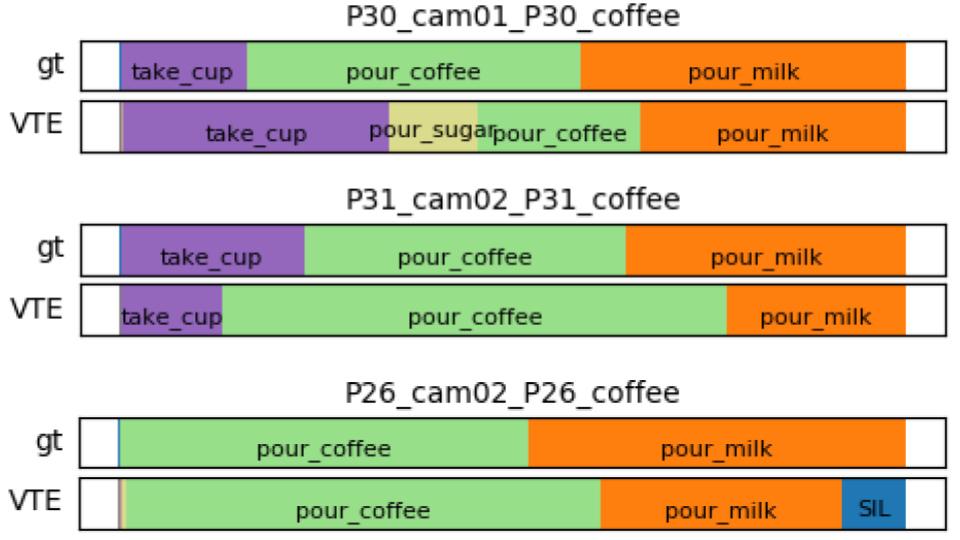}
    \end{subfigure} 
    \begin{subfigure}{0.33\textwidth}
    \includegraphics[width=\linewidth]{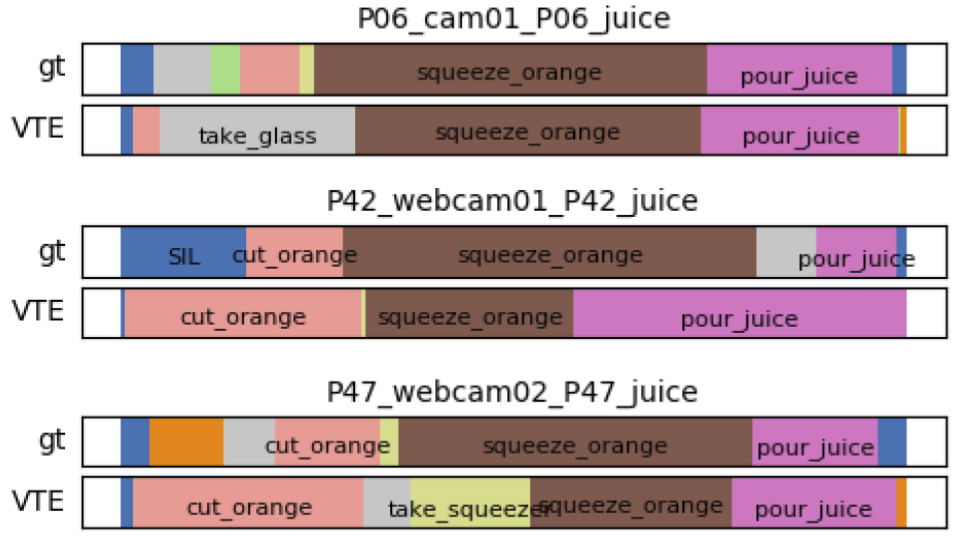}
    \end{subfigure} 
    \begin{subfigure}{0.33\textwidth}
    \includegraphics[width=\linewidth]{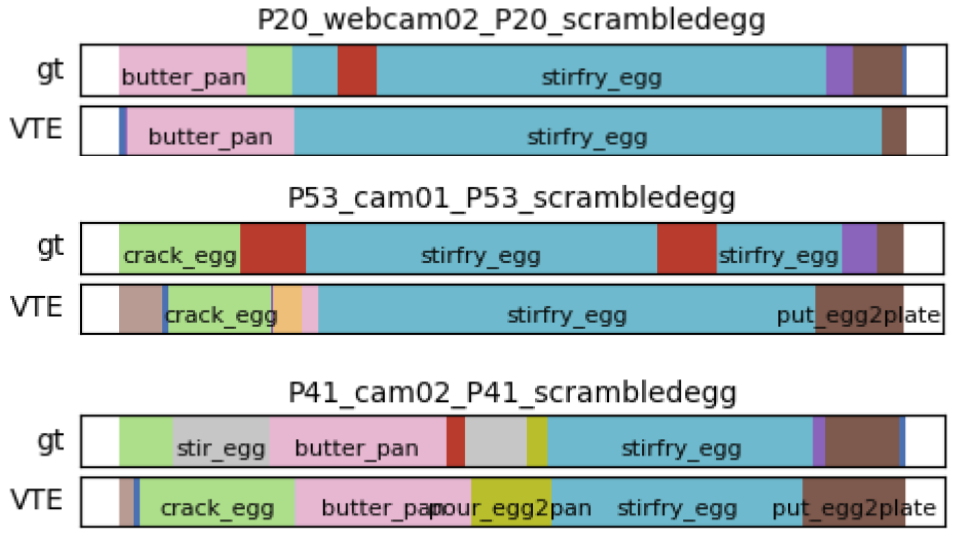}
    \end{subfigure} 
\caption{Segmentation examples from our approach on videos from the Breakfast Actions dataset. We omit some of the ground-truth labels in the visualizations to facilitate reading the labels of the more prominent segments.\vspace{-0.75em}}
\label{fig:examples}
\end{figure*}

\textbf{The Breakfast Actions (BF) dataset} contains 70 hours of cooking activities of varying complexity. It contains 10 different cooking tasks (with about 170 videos per task), which can be further split into 48 sub-activities. The length of each video is highly dependent on the type of task, ranging from 30 seconds to a few minutes. The videos are recorded in different real-life environments with 52 people performing each of the 10 different actions. They have a fixed viewpoint through all activities for each person, which leads to a high intra-class and low inter-class variance.  

\textbf{The INRIA YouTube Instructional Videos (YTI) dataset} contains five tasks of different instructional domains: ``making coffee", ``changing a car tire", ``CPR", ``jumping a car", and ``potting a plant" (with about 30 videos per task), which can be divided into 47 sub-activities. As opposed to Breakfast Actions, the videos might have been edited and include shot boundaries, as well as different and changing viewpoints or zooming. The videos in this dataset are in average longer than the videos in Breakfast Actions, however, there is a significant presence of background frames whereas the Breakfast Action actions are densely labeled without intermediate background classes.

\textbf{The 50 Salads (50S) dataset} contains over 4.5 hours of video captures of different actors preparing 2 kinds of mixed salads (with 25 videos for each type of salad). While similar in nature to the data present in the Breakfast dataset, the videos in 50 Salads tend to be longer and contain more sub-activities. These videos have an average length of 10k frames and contain complex interactions between hands, utensils, and ingredients. In addition to this, different actions in this dataset often contain similar motions, differing only on the ingredient or utensil used and can be further classified into three stages: preparing the dressing, cutting and mixing ingredients, and serving. As such we have similar actions like ``adding oil" or ``adding vinegar" and ``cutting tomato" or ``cutting cucumber" which follow similar hand motions (but differ in the interacting objects) and can happen at interchangeable times.

\subsection{Training} \vspace{-0.75em}

Following the two-stage training protocol described in Section~\ref{systemdescription}, we first train the visual and temporal models independent from each other. The visual embedding model is trained for 160 (for the BF Dataset) and 140 epochs (for the YTI and 50S datasets) while the temporal embedding model is trained for 10, 20, and 30 epochs (for 50S, YTI, and BF respectively). 

During our second training stage, we alternate the training of the visual and temporal models: we train the visual model for 40 consecutive epochs (in which the temporal model is frozen) and then train the temporal discriminator for 5 epochs (where the visual embedding model remains constant. We run these alternating periods for 60 (for the BF dataset) and 120 epochs (for the YTI and 50S datasets).

\subsection{Mapping} \label{evaluation:metrics} \vspace{-0.75em}

To evaluate the segmentation returned by our unsupervised approach, we need to map the segmented clusters of sub-activities to the ground-truth sub-activities related to each specific task. For this, we use Hungarian matching to provide a one-to-one mapping that maximizes the similarity between the segmented clusters and the ground-truth sub-activities. We follow the protocol of~\cite{bojanowski2014weakly,sener2018unsupervised,Kukleva_2019_CVPR} and compute the Hungarian matching over all the videos of one activity. Note that this is different from the Hungarian matching for every single video as used by~\cite{Aakur_2019_CVPR}, which optimizes the matching for every single video and usually leads to higher accuracy, but also allows clusters to change their label from one video to another.

We evaluate the accuracy of our method using two common evaluation metrics used in~\cite{sener2018unsupervised,Kukleva_2019_CVPR,Aakur_2019_CVPR}, 1) \textit{Mean Over Frames (MoF)} to indicate the percentage of frames in the segmentation that were correctly labeled over all the frames of videos assigned to a given task for the Breakfast Actions and 50 Salads Datasets, and 2) \textit{F1} score for the INRIA YouTube Instructional Videos (YTI) Dataset to compare with other state-of-the-art approaches.

\vspace{-0.75em}
\section{Results and Analysis} \vspace{-0.5em}

We compare the performance of our two-stage visual embedding pipeline to state-of-the-art approaches (including weakly-supervised and fully supervised setups) on three challenging action segmentation datasets. As can be seen in Figure~\ref{fig:examples} our approach is able to maintain the logical ordering of sub-activities and a good estimate on their start and duration in actions with different levels of complexity. 

\textbf{Breakfast Actions Dataset:} Our method is able to outperform the state-of-the-art on the unsupervised learning methods applied to this dataset (see Table~\ref{tab:breakfast}) and performs competitively against weakly-supervised approaches. It is important to note that our cluster-to ground-truth mapping and evaluation is done in a global manner, that is, we use all the predicted labels and ground-truth for all the videos of a given task to do the Hungarian matching, and then evaluate calculating the MoF using the count of all the true predictions on the whole dataset. On the other hand, approaches such as LSTM+AL~\cite{Aakur_2019_CVPR} employ a per-video (local) cluster to ground-truth mapping, which might account for the difference in the performance of the two approaches. For comparison purposes, when applying a video-based Hungarian matching our method obtains a MoF of $52.25\%$.

\begin{table}[t]
\centering
\subimport{./tables/}{breakfast.tex} \vspace{-0.5em}
\caption{Segmentation results on the Breakfast Action dataset compared to supervised, weakly supervised, and unsupervised state-of-the-art methods. (*denotes results with video-based Hungarian matching).\vspace{-0.75em}}
\label{tab:breakfast}
\end{table}

\begin{table}[t]
\centering
\subimport{./tables/}{youtube.tex} \vspace{-0.5em}
\caption{Segmentation results on the YTI dataset compared to unsupervised state-of-the-art methods. (*denotes results with video-based Hungarian matching).\vspace{-0.75em}}
\label{tab:youtube}
\end{table}

\begin{figure}
    \centering
    \includegraphics[width=0.95\linewidth]{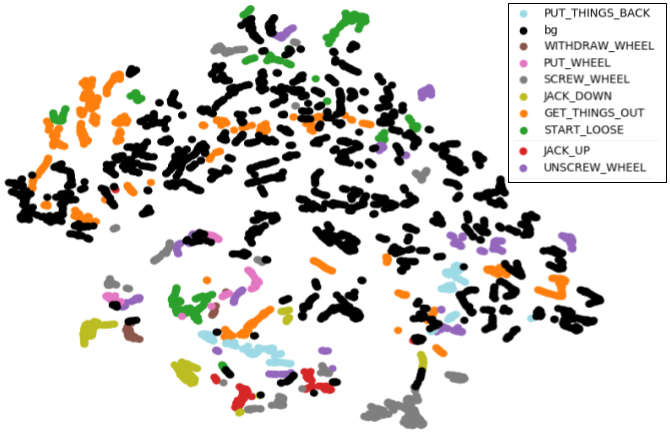} \vspace{-0.5em}
    \caption{t-SNE\cite{wattenberg2016how} Visualization of the cluster distribution on a video from the YTI Dataset. Notice the heavy presence of background (bg) frames. \vspace{-0.75em}}
    \label{fig:yti_hst}
\end{figure}

\textbf{INRIA YouTube Instructional Videos (YTI):} The YTI Dataset is particularly challenging as the majority of its frames consist of ``background" information (see Figure~\ref{fig:yti_hst}). That is, most of the frames in this dataset do not provide any relevant visual cues related to the action we are segmenting. In line with other approaches, our evaluation on this dataset excludes any frames our method assigned to the label background so as to avoid any bias towards the over-segmentation of this sub-activity. Despite the reduced amount of valuable visual cues we can gather from single frames in this data, our method outperforms other global-based Hungarian matching approaches (see Table~\ref{tab:youtube}). We observed a considerable improvement when using larger frame prediction steps (predicting further into the future) for this dataset, which probably helped our visual embedding model to discriminate the minutiae associated with irrelevant frames.

\begin{table}[]
\centering
\subimport{./tables/}{salads.tex} \vspace{-0.5em}
\caption{MoF Segmentation results on the 50 Salads dataset at mid and eval granularity compared to state-of-the-art methods. (*denotes results with video-based Hungarian matching) Blank fields were left for methods that did not report a score for a type of granularity.\vspace{-0.75em}}
\label{tab:salads}
\end{table}

\textbf{50 Salads (50S):} While more similar in nature to the videos in Breakfast Actions, the order of the sub-activities in 50 Salads is not as important as with the previous datasets. Sub-activities belonging to the same category have similar temporal locations (\eg, ``cutting  tomato", and ``cutting cucumber"), and often the order in which they are performed does not have a big impact on the performance of the task. Consequently the benefits we see from following the two training stages (Table~\ref{tab:salads}) are not as significant as for the other two datasets, just using the visual embedding from the first stage seemed to provide fewer distractions (brought on by the weak temporal ordering of the actions in this dataset) and resulted on a better performance (see Table~\ref{tab:steps}).

\subsection{Ablation Studies}\label{sec:ablation}\vspace{-0.75em}
\subsubsection{Future frame prediction task}\vspace{-0.5em}

In order to assess the value of future frame prediction as a good learning task to encode both the visual and temporal cues of the data, we tested the performance of our visual embedding model (Section~\ref{sys:s1:visual}) using a frame prediction step sizes $s$ of $0$, $1$, $3$, and $5$ (see Table~\ref{tab:steps}). When $s=0$ the network would learn to reconstruct the same input frame from the generated encoding (rather than generating a prediction of a future frame), with $s>0$ the network would predict the next $s_{th}$ frame in the video sequence following the input frame (we should be able to generate a given frame $t$ from the encoding of a previously seen frame at time $t-s$). 

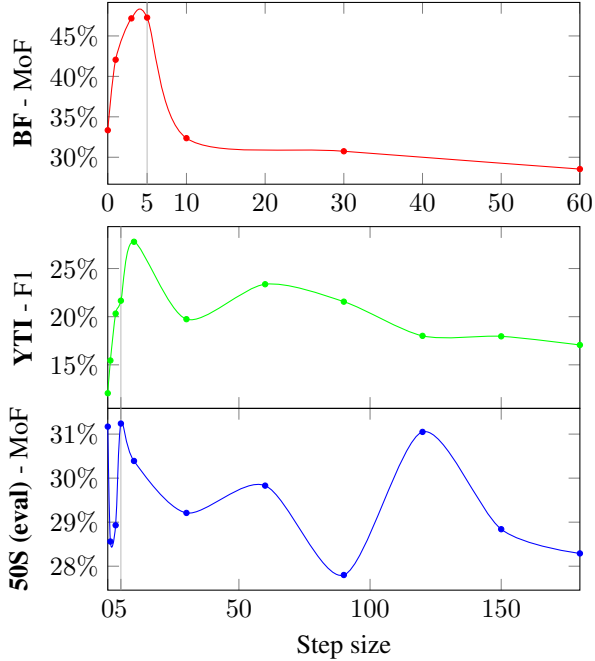
\begin{figure}[t]
    \centering
    \subimport{./figures/}{step_growth.tex}\vspace{-0.5em}
    \caption{U-Net Embedding Performance when predicting longer step sizes ($s$). We do not test $s>60$ for BF as some of the videos in the dataset have  as few as 180 frames.\vspace{-1em}}
    \label{fig:step_growth}
\end{figure}

We observed a distinctive improvement for both the BF and YTI Datasets as the step size increases, with the lowest performance obtained by just reconstructing the original input frame ($s=0$). The strong correlation between temporal and visual cues in these datasets allowed us to extract a more significant encoding using our approach. However, it is important to note that, while our approach benefits from the future frame prediction task, when testing overly large step sizes (see Fig.~\ref{fig:step_growth}) the performance degrades as the step size increases. Overly large step sizes could exacerbate the erosion of shorter sub-activities from the temporal segmentation as well as disrupt the learning of the logical ordering of the sub-activities in a given action. 

This was not the case for the 50 Salads dataset, while overall the U-Net with the largest step size obtained the best performance, the further enforcement of the temporal encoding brought on by the second stage of our training pipeline proved to be detrimental as the temporal cues are not that significant in this type of data.

\begin{table}[t]
\centering
\subimport{./tables/}{steps.tex}
\caption{U-Net Learning task impact: Given an input frame captured at a time $t$ the U-Net was trained to estimate the frame $s$ steps after the input frame. MoF is reported for BF and 50 Salads, F1 score is reported for YTI.\vspace{-0.75em}}
\label{tab:steps}
\end{table}

\vspace{-1em}
\subsubsection{Feature Embedding}\vspace{-0.5em}


The purpose of our embedding algorithm is to encode valuable visual and temporal information derived from our input features, to evaluate the impact of such embedding, we first analyze the performance of the segmentation pipeline when omitting the feature embedding step. For this, we skipped the embedding process and directly clustered the reduced Fisher Vectors following the process described in Sec.~\ref{sec:clusndecod}. We observed a significant decrease in the performance obtained for all three datasets, with a 13.5\% F1 score for the INRIA Youtube Instructional Videos (YTI) Dataset, and an MoF score of 28.13\% and 29.07\% for the Breakfast Actions (BF) and 50 Salads (50S) Datasets respectively.

As described in Sec.~\ref{systemdescription}, we train our visual (U-Net) and temporal (MLP) components in two stages to obtain a joint visual and temporal embedding (extracted from the visual model after the second stage). We now evaluate the impact of different types of embedding (visual, temporal, or joint) during both stages of our training. Table~\ref{tab:embedding} shows the result using the embedding of the visual and temporal components, the U-Net and the MLP respectively. At stage 1 both components were trained in a disjoint fashion, following the losses described in Equations~\ref{f:visual_loss} (U-Net) and ~\ref{f:temporal_loss} (MLP). At stage 2 the loss of both models is updated so that they are trained jointly (alternating the training of the U-Net and MLP for 40 and 5 epochs respectively) following Equations~\ref{f:unet_loss} (U-Net) and \ref{f:mlp_loss} (MLP). We used the output of the U-Net's encoder as our embedding for the U-Net, and the output of the inner layer of the MLP as its embedding. Overall the visual embedding of the U-Net after the first stage already led to an increased performance of the whole system. Nevertheless, further enforcing a stronger temporal embedding through stage 2 provided the best performance for the Breakfast Actions Dataset.

\begin{table}[t]
\centering
\subimport{./tables/}{embeddings.tex} \vspace{-0.5em}
\caption{Analysis of the feature embedding sources on the Breakfast Actions dataset at both stages of training. Stage 1 Models were trained independently of each other.\vspace{-1em}}
\label{tab:embedding}
\end{table}

\vspace{-0.75em}
\section{Discussion}

Various approaches have been proposed for learning visual representations without labels~\cite{Jing2019survey, Kukleva_2019_CVPR, Aakur_2019_CVPR, sener2018unsupervised}. Particularly, in the case of learning video representations in an unsupervised way, the temporal properties of the data have been commonly employed to understand the structure of the observed visual features and the rules they follow while they change over time. Following the idea that given a set of videos all capturing the same activity it should be possible to identify temporal segments with similar sub-actions across all videos, we have studied the suitability of a joint visual-temporal approach to learn to identify different action segments from unlabeled video data.

The results of our experiments have shown that a combination of predictive visual and temporal learning tasks encourages the encoding of valuable information for temporal segmentation that might have been otherwise disregarded by a solely temporal approach. The segmentations obtained with our method maintain the logical ordering of the analyzed tasks and are able to produce coherent segments that have led us to significant improvements over previous approaches that have exclusively used temporal embeddings. 

We have observed that incorporating the visual cues that might be present in a frame into our joint approach has shown promise in the area of temporal segmentation, particularly for cases in which the sub-activities have either 1) \textit{distinct visual} appearances (e.g., ``cracking egg'' vs. ``buttering pan''), and/or 2) a \textit{strong temporal coherence} (e.g., the ``making coffee'' sub-activities ``pouring coffee'' and ``pouring milk'' would have similar visual aspects, however, in most cases the actors in the videos pour the coffee (into the mug) first before adding milk to it), as was the case for datasets like Breakfast Actions and INRIA  Youtube Instructions (YTI)). However, our approach struggles with data that has a weak temporal coherence and low visual variance (like the data in 50 Salads). We suggest as a future research direction incorporating an evaluation of the temporal coherence of sub-activities, both to address this issue and to incorporate the segmentation of repetitive actions (which lead to inconsistent results in their temporal location).

{\small
\bibliographystyle{ieee_fullname}
\bibliography{egbib}
}
\end{document}

%% file: tables/breakfast.tex
\begin{tabular}{|l|l|l|}
\hline 
	 \textbf{Supervision} & \textbf{Approach}	& \textbf{MoF} 	\\ \hline \hline
\multirow{6}{*}{Full}	
	& TCFPN	\cite{ding2018weakly}	        & 52.0\%	\\
	& HTK \cite{kuehne2016end}	        & 56.3\%	\\ 
	& GRU \cite{richard2017weakly}	        & 60.6\%	\\
	& MS-TCNN++ \cite{li2020mstcn} & 67.6\% \\
    & Local SSTDA~\cite{chen2020action}        & 70.2\%	\\ 
    & SSTDA\cite{lea2016segmental}  & \textbf{70.3\%}	\\
    \hline
\multirow{4}{*}{Weak} & Fine2Coarse \cite{richard2016temporal}    & 33.3\%	\\
    & GRU \cite{richard2017weakly}  & 36.7\% \\
	& TCFPN+ISBA	\cite{ding2018weakly}	& 38.4\%	\\ 
	& NN-Viterbi \cite{richard2018nnviterbi}    & 43\%	\\ 
	& D3TW \cite{chang2019d3tw}	                                    & 45.7\%	\\ 
	& CDFL \cite{li2019weakly}                                     & \textbf{50.2\%} \\
	\hline
\multirow{4}{*}{Unsupervised}	& GMM \cite{sener2018unsupervised}	& 34.6\%	\\
	& CTE-MLP \cite{Kukleva_2019_CVPR}          & 41.8\%           	\\
	& (LSTM+AL \cite{Aakur_2019_CVPR})	& (42.9\%*)	\\ 
	\cline{2-3}
	& Ours                         &	\textbf{48.08\%}	\\ 
	\hline
\end{tabular}

%% file: tables/youtube.tex
\begin{tabular}{|l|l|} 
\hline
\textbf{Approach} & \textbf{F1} \\ 
\hline 
\hline
GMM \cite{sener2018unsupervised}  & 27.0\%\\
CTE-MLP \cite{Kukleva_2019_CVPR}  & 28.3\%  \\
(LSTM+AL \cite{Aakur_2019_CVPR})  & (39.7\%*)  \\
 \hline
Ours & \textbf{29.86\%} \\ 
\hline
\end{tabular}

%% file: tables/salads.tex
\begin{tabular}{|l|l|l|l|}
\hline 
	 \textbf{Supervision} & \textbf{Approach}	& \textbf{Eval} & \textbf{Mid} 	\\ \hline \hline
\multirow{6}{*}{Full}	
    & S-CNN \cite{lea2016segmental}	        & 68.0\% & 54.9\%	\\ 
    & ST-CNN\cite{lea2016segmental}      & 72.00\% & 58.06\%	\\
	& ED-TCN \cite{lea2017temporal}	        & 73.4\% & 64.7\%	\\
	& MS-TCNN++ \cite{li2020mstcn} & -- & 83.7\%\\
    & Local SSTDA~\cite{chen2020action}	        & -- & 82.8\%	\\ 
    & SSTDA\cite{lea2016segmental}      & -- & \textbf{83.8\%}	\\
    \hline
\multirow{3}{*}{Unsupervised}
	& CTE-MLP \cite{Kukleva_2019_CVPR}          & \textbf{35.5\%} & \textbf{30.2\%} \\
	& (LSTM+AL \cite{Aakur_2019_CVPR})	 & (60.6\%)* & - \\
	\cline{2-4}
	& Ours & 30.59\% & 24.19\%	\\ 
	\hline
\end{tabular}

%% file: tables/steps.tex
\begin{tabular}{|l|l|l|l|}
\hline 
\textbf{Step size} & \textbf{BF} & \textbf{YTI}  & \textbf{50S (eval)}  \\ \hline \hline
U-Net (s=0)   & 33.34\% & 12.16\% & 31.17\%\\ 
U-Net (s=1)   & 42.06\% & 13.51\% & 28.56\%\\ 
U-Net (s=3)   & 47.17\% & 21.32\% & 28.93\%\\ 
U-Net (s=5)   & \textbf{47.27\%} & \textbf{25.72\%} & \textbf{31.24\%}\\ \hline
\end{tabular}

%% file: figures/step_growth.tex
\begin{tikzpicture}




\begin{groupplot}[
        group style={
            group size=1 by 1,
            x descriptions at=edge bottom,
            vertical sep=0pt,
        },
        xmin = 0, xmax = 60, 
        height=4cm,
        width=0.45*\textwidth,
        extra x ticks={5},
        extra x tick style={grid=major},
        yticklabel={$\pgfmathprintnumber{\tick}\%$}
    ]
    \nextgroupplot[
        ylabel={\textbf{BF} - MoF},
    ]
        \addplot [red, smooth, mark=*, mark size=1pt]  coordinates { 
            (0, 33.34)
            (1, 42.06)
            (3, 47.17)
            (5, 47.27)
            (10, 32.36)
            (30, 30.74)
            (60, 28.54)
        };
\end{groupplot}

\begin{groupplot}[
        group style={
            group size=1 by 2,
            x descriptions at=edge bottom,
            vertical sep=0pt,
            xticklabels at=edge bottom,
        },
        xlabel=Step size,
        xmin = 0, xmax = 180, 
        height=4cm,
        width=0.45*\textwidth,
        extra x ticks={5},
        extra x tick style={grid=major},
        yticklabel={$\pgfmathprintnumber{\tick}\%$},
        xtick align=inside,
    ]

    \nextgroupplot[
        ylabel={\textbf{YTI} - F1},
        at={(rel axis cs:0,-0.15)},
        anchor=north west,
        xticklabel style={color=white},
    ]
        \addplot [green, smooth, mark=*, mark size=1pt] coordinates 
        { 
            (0, 12.05)
            (1, 15.45)
            (3, 20.32)
            (5, 21.66)
            (10, 27.78)
            (30, 19.74)
            (60, 23.37)
            (90, 21.56)
            (120, 18.01)
            (150, 17.96)
            (180, 17.06)
        };
    
    \nextgroupplot[
        ylabel={\textbf{50S (eval)} - MoF}
    ]
        \addplot [blue, smooth, mark=*, mark size=1pt] coordinates 
        { 
            (0, 31.17)
            (1, 28.56)
            (3, 28.93)
            (5, 31.24)
            (10, 30.39)
            (30, 29.21)
            (60, 29.83)
            (90, 27.80)
            (120, 31.05)
            (150, 28.84)
            (180, 28.29)
        };
    \end{groupplot}

\end{tikzpicture}

%% file: tables/embeddings.tex
\begin{tabular}{|l|l|l|}
\hline
\textbf{Training} & \textbf{Embedding Source} & \textbf{MoF}       \\ \hline\hline
-- & Fisher Vector (no embedding) & 28.13\% \\ \hline\hline
\multirow{2}{*}{Stage 1} & U-Net              & 47.27\%           \\
 & MLP                & 40.91\%          \\\hline
\multirow{2}{*}{Stage 2} & U-Net   & \textbf{48.08}\%           \\
 & MLP     & 39.04\% \\ \hline
\end{tabular}

%% file: main.bbl
\begin{thebibliography}{10}\itemsep=-1pt

\bibitem{Aakur_2019_CVPR}
Sathyanarayanan~N. Aakur and Sudeep Sarkar.
\newblock A perceptual prediction framework for self supervised event
  segmentation.
\newblock In {\em CVPR}, June 2019.

\bibitem{Alayrac_2016_CVPR}
J. {Alayrac}, P. {Bojanowski}, N. {Agrawal}, J. {Sivic}, I. {Laptev}, and S.
  {Lacoste-Julien}.
\newblock Unsupervised learning from narrated instruction videos.
\newblock In {\em CVPR}, 2016.

\bibitem{badrinarayanan2017segnet}
Vijay Badrinarayanan, Alex Kendall, and Roberto Cipolla.
\newblock Segnet: A deep convolutional encoder-decoder architecture for image
  segmentation.
\newblock {\em PAMI}, 39(12):2481--2495, 2017.

\bibitem{bojanowski2014weakly}
Piotr Bojanowski, R{\'e}mi Lajugie, Francis Bach, Ivan Laptev, Jean Ponce,
  Cordelia Schmid, and Josef Sivic.
\newblock Weakly supervised action labeling in videos under ordering
  constraints.
\newblock In {\em ECCV}, 2014.

\bibitem{chang2019d3tw}
Chien-Yi Chang, De-An Huang, Yanan Sui, Li Fei-Fei, and Juan~Carlos Niebles.
\newblock D3tw: Discriminative differentiable dynamic time warping for weakly
  supervised action alignment and segmentation.
\newblock In {\em CVPR}, pages 3546--3555, 2019.

\bibitem{chen2020action}
Min-Hung Chen, Baopu Li, Yingze Bao, Ghassan AlRegib, and Zsolt Kira.
\newblock Action segmentation with joint self-supervised temporal domain
  adaptation, 2020.

\bibitem{NIPS2017_7028}
Emily~L Denton and vighnesh Birodkar.
\newblock Unsupervised learning of disentangled representations from video.
\newblock In I. Guyon, U.~V. Luxburg, S. Bengio, H. Wallach, R. Fergus, S.
  Vishwanathan, and R. Garnett, editors, {\em NIPS}, pages 4414--4423. Curran
  Associates, Inc., 2017.

\bibitem{ding2018weakly}
Li Ding and Chenliang Xu.
\newblock Weakly-supervised action segmentation with iterative soft boundary
  assignment.
\newblock In {\em CVPR}, 2018.

\bibitem{goodfellow2014generative}
Ian Goodfellow, Jean Pouget-Abadie, Mehdi Mirza, Bing Xu, David Warde-Farley,
  Sherjil Ozair, Aaron Courville, and Yoshua Bengio.
\newblock Generative adversarial nets.
\newblock In {\em NIPS}, pages 2672--2680, 2014.

\bibitem{huang2016connectionist}
De-An Huang, Li Fei-Fei, and Juan~Carlos Niebles.
\newblock Connectionist temporal modeling for weakly supervised action
  labeling.
\newblock In {\em ECCV}, pages 137--153. Springer, 2016.

\bibitem{isola2017image}
Phillip Isola, Jun-Yan Zhu, Tinghui Zhou, and Alexei~A Efros.
\newblock Image-to-image translation with conditional adversarial networks.
\newblock In {\em CVPR}, pages 1125--1134, 2017.

\bibitem{Jing2019survey}
Longlong Jing and Yingli Tian.
\newblock Self-supervised visual feature learning with deep neural networks:
  {A} survey.
\newblock {\em CoRR}, abs/1902.06162, 2019.

\bibitem{Kuehne_2014_CVPR}
H. {Kuehne}, A. {Arslan}, and T. {Serre}.
\newblock The language of actions: Recovering the syntax and semantics of
  goal-directed human activities.
\newblock In {\em CVPR}, 2014.

\bibitem{kuehne2016end}
Hilde Kuehne, Juergen Gall, and Thomas Serre.
\newblock An end-to-end generative framework for video segmentation and
  recognition.
\newblock In {\em WACV}, 2016.

\bibitem{kuehne2019hybird}
H. {Kuehne}, A. {Richard}, and J. {Gall}.
\newblock A hybrid rnn-hmm approach for weakly supervised temporal action
  segmentation.
\newblock {\em IEEE Transactions on Pattern Analysis and Machine Intelligence},
  pages 1--1, 2019.

\bibitem{Kukleva_2019_CVPR}
Anna Kukleva, Hilde Kuehne, Fadime Sener, and Jurgen Gall.
\newblock Unsupervised learning of action classes with continuous temporal
  embedding.
\newblock In {\em CVPR}, 2019.

\bibitem{lea2017temporal}
Colin Lea, Michael~D Flynn, Rene Vidal, Austin Reiter, and Gregory~D Hager.
\newblock Temporal convolutional networks for action segmentation and
  detection.
\newblock In {\em CVPR}, pages 156--165, 2017.

\bibitem{lea2016segmental}
Colin Lea, Austin Reiter, Ren{\'e} Vidal, and Gregory~D Hager.
\newblock Segmental spatiotemporal cnns for fine-grained action segmentation.
\newblock In {\em European Conference on Computer Vision}, pages 36--52.
  Springer, 2016.

\bibitem{li2019weakly}
Jun Li, Peng Lei, and Sinisa Todorovic.
\newblock Weakly supervised energy-based learning for action segmentation.
\newblock In {\em ICCV}, pages 6243--6251, 2019.

\bibitem{li2020mstcn}
Shijie Li, Yazan~Abu Farha, Yun Liu, Ming-Ming Cheng, and Juergen Gall.
\newblock Ms-tcn++: Multi-stage temporal convolutional network for action
  segmentation, 2020.

\bibitem{long2015fully}
Jonathan Long, Evan Shelhamer, and Trevor Darrell.
\newblock Fully convolutional networks for semantic segmentation.
\newblock In {\em CVPR}, pages 3431--3440, 2015.

\bibitem{DBLP:journals/corr/LotterKC15}
William Lotter, Gabriel Kreiman, and David~D. Cox.
\newblock Unsupervised learning of visual structure using predictive generative
  networks.
\newblock {\em ICLR}, abs/1511.06380, 2015.

\bibitem{DBLP:journals/corr/LotterKC16}
William Lotter, Gabriel Kreiman, and David~D. Cox.
\newblock Deep predictive coding networks for video prediction and unsupervised
  learning.
\newblock {\em ICLR}, abs/1605.08104, 2016.

\bibitem{malmaud2015s}
Jonathan Malmaud, Jonathan Huang, Vivek Rathod, Nick Johnston, Andrew
  Rabinovich, and Kevin Murphy.
\newblock What's cookin'? interpreting cooking videos using text, speech and
  vision.
\newblock {\em NAACL}, 2015.

\bibitem{mathieu2015deep}
Michael Mathieu, Camille Couprie, and Yann LeCun.
\newblock Deep multi-scale video prediction beyond mean square error.
\newblock {\em ICLR}, 2015.

\bibitem{mirza2014conditional}
Mehdi Mirza and Simon Osindero.
\newblock Conditional generative adversarial nets.
\newblock {\em NIPS}, 2014.

\bibitem{MisraZH16Shuffle}
Ishan Misra, C.~Lawrence Zitnick, and Martial Hebert.
\newblock Shuffle and learn: Unsupervised learning using temporal order
  verification.
\newblock In {\em ECCV}, 2016.

\bibitem{peng2014action}
Xiaojiang Peng, Changqing Zou, Yu Qiao, and Qiang Peng.
\newblock Action recognition with stacked fisher vectors.
\newblock In {\em European Conference on Computer Vision}, pages 581--595.
  Springer, 2014.

\bibitem{richard2016temporal}
Alexander Richard and Juergen Gall.
\newblock Temporal action detection using a statistical language model.
\newblock In {\em CVPR}, pages 3131--3140, 2016.

\bibitem{richard2017weakly}
Alexander Richard, Hilde Kuehne, and Juergen Gall.
\newblock Weakly supervised action learning with rnn based fine-to-coarse
  modeling.
\newblock In {\em CVPR}, 2017.

\bibitem{richard2018nnviterbi}
Alexander Richard, Hilde Kuehne, Ahsan Iqbal, and Juergen Gall.
\newblock Neuralnetwork-viterbi: A framework for weakly supervised video
  learning.
\newblock In {\em CVPR}, 2018.

\bibitem{DBLP:journals/corr/RonnebergerFB15}
Olaf Ronneberger, Philipp Fischer, and Thomas Brox.
\newblock U-net: Convolutional networks for biomedical image segmentation.
\newblock {\em MICCAI}, abs/1505.04597, 2015.

\bibitem{sener2018unsupervised}
Fadime Sener and Angela Yao.
\newblock Unsupervised learning and segmentation of complex activities from
  video.
\newblock In {\em CVPR}, pages 8368--8376, 2018.

\bibitem{DBLP:journals/corr/SenerZWSS16}
Ozan Sener, Amir~Roshan Zamir, Chenxia Wu, Silvio Savarese, and Ashutosh
  Saxena.
\newblock Unsupervised semantic action discovery from video collections.
\newblock {\em CoRR}, abs/1605.03324, 2016.

\bibitem{srivastava2015Unsupervised}
Nitish Srivastava, Elman Mansimov, and Ruslan Salakhudinov.
\newblock Unsupervised learning of video representations using lstms.
\newblock In {\em ICML}, 2015.

\bibitem{stein2013combining}
Sebastian Stein and Stephen~J McKenna.
\newblock Combining embedded accelerometers with computer vision for
  recognizing food preparation activities.
\newblock In {\em Proceedings of the 2013 ACM international joint conference on
  Pervasive and ubiquitous computing}, pages 729--738, 2013.

\bibitem{wattenberg2016how}
Martin Wattenberg, Fernanda Viégas, and Ian Johnson.
\newblock How to use t-sne effectively.
\newblock {\em Distill}, 2016.

\bibitem{DBLP:journals/corr/WhitneyCKT16}
William~F. Whitney, Michael Chang, Tejas~D. Kulkarni, and Joshua~B. Tenenbaum.
\newblock Understanding visual concepts with continuation learning.
\newblock {\em ICLR}, abs/1602.06822, 2016.

\bibitem{zacks2007event}
Jeffrey~M. Zacks and Khena~M. Swallow.
\newblock Event segmentation.
\newblock {\em Current Directions in Psychological Science}, 16(2):80--84,
  2007.

\bibitem{zhu2017toward}
Jun-Yan Zhu, Richard Zhang, Deepak Pathak, Trevor Darrell, Alexei~A Efros,
  Oliver Wang, and Eli Shechtman.
\newblock Toward multimodal image-to-image translation.
\newblock In {\em NIPS}, pages 465--476, 2017.

\end{thebibliography}
